\documentclass[conference]{IEEEtran}
\usepackage{bm}
\usepackage{comment}

\ifCLASSINFOpdf
  \usepackage[pdftex]{graphicx}
\else
  \usepackage[dvips]{graphicx}
\fi
\begin{document}
%
\title{Gradient-Based Low-Light Image Enhancement}

\author{\IEEEauthorblockN{Masayuki Tanaka}
\IEEEauthorblockA{AIST \\
Tokyo Institute of Technology}
\and
\IEEEauthorblockN{Takashi Shibata}
\IEEEauthorblockA{NEC Corporation}
\and
\IEEEauthorblockN{Masatoshi Okutomi}
\IEEEauthorblockA{Tokyo Institute of Technology}}


%


\maketitle

\begin{abstract}
A low-light image enhancement is a highly demanded image processing technique, especially for consumer digital cameras and cameras on mobile phones. 
In this paper, a gradient-based low-light image enhancement algorithm is proposed.
The key is to enhance the gradients of dark region, because the gradients are more sensitive for human visual system than absolute values.
In addition, we involve the intensity-range constraints for the image integration.
By using the intensity-range constraints, we can integrate the output image with enhanced gradients preserving the given gradient information while enforcing the intensity range of the output image within a certain intensity range.
Experiments demonstrate that the proposed gradient-based low-light image enhancement can effectively enhance the low-light images.
\end{abstract}



%
\IEEEpeerreviewmaketitle

\section{Introduction}
Low-light image enhancement is a highly demanded image processing technique, especially for consumer digital cameras and cameras on mobile phones. 
A histogram equalization~\cite{Russ_HistEq} and a non-linear tone curve adjustment are widely used simple algorithms. Recently, several sophisticated low-light image enhancement algorithms have been proposed~\cite{Yuan_Exposure,Guo_LIME,Li_LowLight}. 

Those algorithms are basically intensity based correction algorithms. A gradient-based image processing frameworks have been proposed as new image processing approaches~\cite{Agrawal_GradientCourse,Bhat_GradientShop,Shibata_IntensityRange,Tanaka_SiggraphPoster,Shibata_EI}. 
The gradient-based image processing is becoming a powerful tool because the gradient is more sensitive for the human visual system (HVS) than the absolute value. 

Then, in this paper, we propose a gradient-based low-light image enhancement. The gradients of a dark region tend to be small. 
The problem comes from those small values of gradient, because those small values of gradients are difficult for the HVS to recognize. 
Then, the basic idea of the proposed algorithm is to simply enhance the gradient of the dark or the low-light region. 
In the proposed algorithm, the gradients of the input image are firstly extracted. Then, the gradients of the dark region are enhanced. 
The output images are generated by integrate the enhanced gradients.

One of advantage of the gradient-based low-light image enhancement is that we can simultaneously apply the gradient-based filtering such as edge sharpening and smoothing. On the other hand, the gradient-based image processing has drawback such as the intensity range of the output image is unknown before the integration or the output image generation. Therefore, in many existing gradient-based image processing, the intensity clipping or the intensity rescaling are implicitly applied after the image integration. The intensity clipping yields a saturation and the intensity rescaling leads that the enhanced gradients are weaken. In order to overcome those problems, we involve the intensity-range constraint for the image integration~\cite{Shibata_IntensityRange}. The image integration with the intensity-range constraint can preserve the gradient while enforcing the intensity range of the output image within given specific range.

The experiments demonstrate that the proposed gradient-based low-light image enhancement can effectively enhance the low-light region or the dark region. 

\section{Proposed Gradient-Based Low-Light Image Enhancement}
Here, we focus on an image processing of a gray image.
For the RGB color image, the color image is firstly converted into the luminance-chrominance color space.
Then, after luminance component is processed, the chrominance components are added to produce the processed RGB image.

\begin{figure*}
 \centering
 \includegraphics[width=0.7\linewidth]{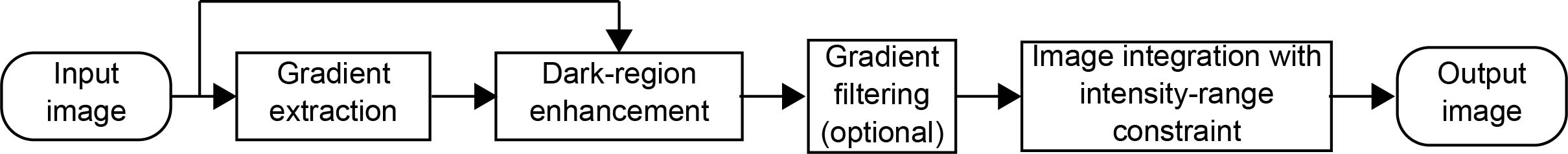}
 \vspace*{-1mm}
 \caption{Image processing pipeline of proposed gradient-based low-light image enhancement.}
 \label{fig:Pipeline}
 \vspace*{-2.75mm}
\end{figure*}

Figure~\ref{fig:Pipeline} shows an overall image processing pipeline of the proposed gradient-based low-light image enhancement.
First, the pixel intensity differences of the horizontal and the vertical adjacent pixels of the input image are calculated as an approximation of the gradients. 
The calculated gradients of the dark region are enhanced as follows:
\begin{eqnarray}
 q_h({\bm x}) &=& f_h({\bm x}) \cdot L(f({\bm x}); \beta, \tau) \,,
 \label{eq:q_h} \\
 q_v({\bm x}) &=& f_v({\bm x}) \cdot L(f({\bm x}); \beta, \tau)  \,,
 \label{eq:q_v}
\end{eqnarray}
where 
${\bm x}$ is the pixel position,
$f({\bm x})$ is the intensity of the input image,
$f_h({\bm x})$ and $f_v({\bm x})$ is the horizontal and the vertical gradients of the input image,
$q_h({\bm x})$ and $q_v({\bm x})$ is the horizontal and the vertical enhanced gradients, 
and
$L(f({\bm x}); \beta, \tau)$ is a enhancement function,
respectively.
In this paper, the enhancement function is designed with two parameters $\beta$ and $\tau$ by
\begin{eqnarray} 
 L(\xi; \beta, \tau) &=&
 \left\{
 \begin{array}{ll}
  \frac{\beta-1}{2\tau^2} \xi^2 - \frac{\beta-1}{\tau} \xi+\beta & (\xi \le \tau) \\
  1 & (\xi > \tau)
 \end{array}
 \right.
 \,,
\end{eqnarray}
where $\xi$ is the pixel intensity.
This enhancement function amplifies the gradient $\beta$ times if associated pixel intensity is zero. 
The amplification ratio for greater than $\tau$ value of intensity is one.
This enhancement function is designed, so that the amplification ratio is smoothly decreasing from intensity value of zero to $\tau$.
In this paper, the parameters $\beta$ and $\tau$ are manually set.
The adaptive design of those parameters will be included in our future works.

Optionally, we can insert any kind of gradient manipulation before the image integration.
Once we obtain the enhanced gradients, or the desired gradients, $q_h({\bm x})$ and $q_v({\bm x})$, we can generate the output image by integrating those gradients. 
In this paper, we follow the integration process with the intensity-range constraint as in~\cite{Shibata_IntensityRange}.
Those process can preserve the gradient information while enforcing the intensity-range of the output image within the specific range which is 0 to 255 for 8-bit image.
For the detailed integration process, please refer the paper~\cite{Shibata_IntensityRange}.

\section{Experiments}
\begin{figure}[t]
 \newlength{\miniwidth}
 \setlength{\miniwidth}{40mm}
 \newlength{\imgwidth}
 \setlength{\imgwidth}{.95\miniwidth}
 \newlength{\vsp}
 \setlength{\vsp}{1mm}

 \centering

 \hfill
 \begin{minipage}[t]{\miniwidth}
  \center
  \includegraphics[width=\imgwidth]{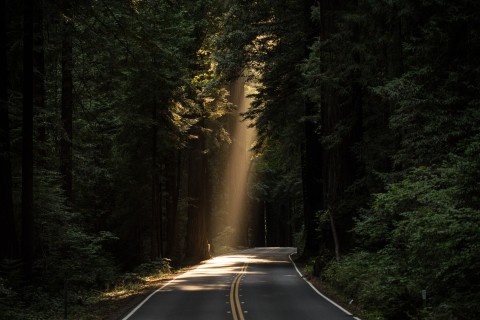}
 \end{minipage}
 \hfill
 \begin{minipage}[t]{\miniwidth}
  \center 
  \includegraphics[width=\imgwidth]{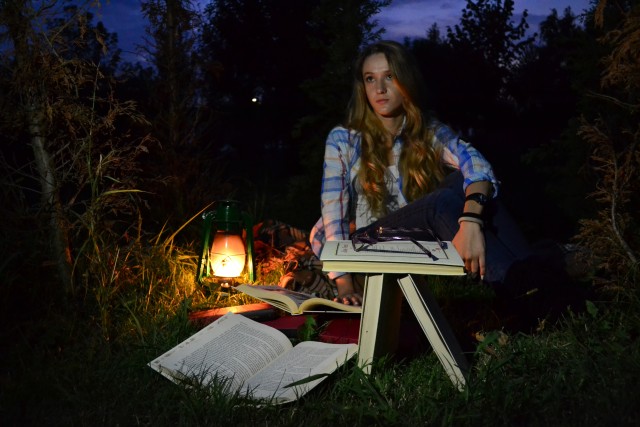} 
 \end{minipage}
 \hfill
 
 \vspace*{-\vsp}
 (a) Input images

 \vspace*{\vsp} \vspace*{\vsp}

\hfill
 \begin{minipage}[t]{\miniwidth}
  \center
  \includegraphics[width=\imgwidth]{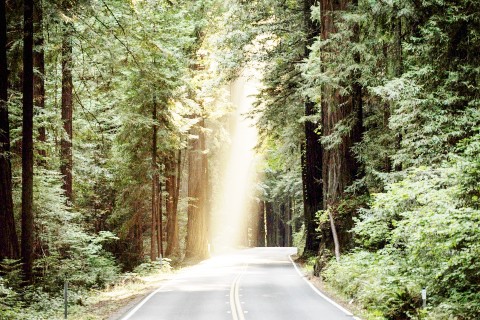}
 \end{minipage}
 \hfill
 \begin{minipage}[t]{\miniwidth}
  \center 
  \includegraphics[width=\imgwidth]{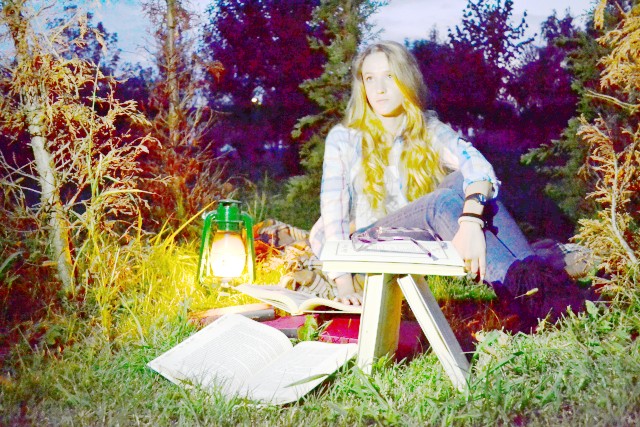} 
 \end{minipage}
 \hfill
 
 \vspace*{-\vsp}
 (b) Histogram equalization

 \vspace*{\vsp} \vspace*{\vsp}

\hfill
 \begin{minipage}[t]{\miniwidth}
  \center
  \includegraphics[width=\imgwidth]{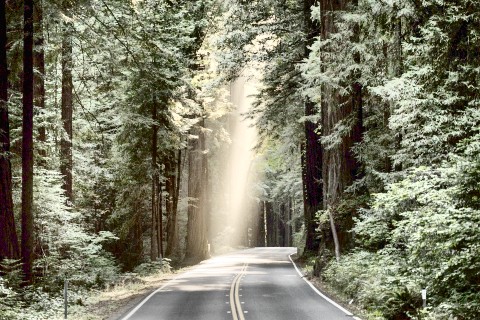}
 \end{minipage}
 \hfill
 \begin{minipage}[t]{\miniwidth}
  \center 
  \includegraphics[width=\imgwidth]{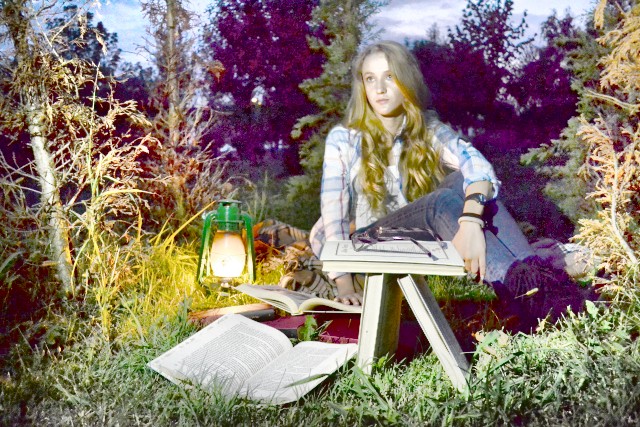} 
 \end{minipage}
 \hfill
 
 \vspace*{-\vsp}
 (c) Matlab HDR tone mapping

 \vspace*{\vsp} \vspace*{\vsp}

\hfill
 \begin{minipage}[t]{\miniwidth}
  \center
  \includegraphics[width=\imgwidth]{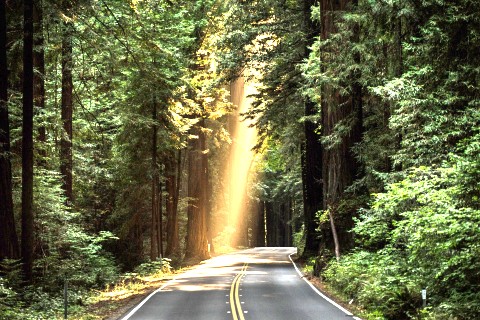}
 \end{minipage}
 \hfill
 \begin{minipage}[t]{\miniwidth}
  \center 
  \includegraphics[width=\imgwidth]{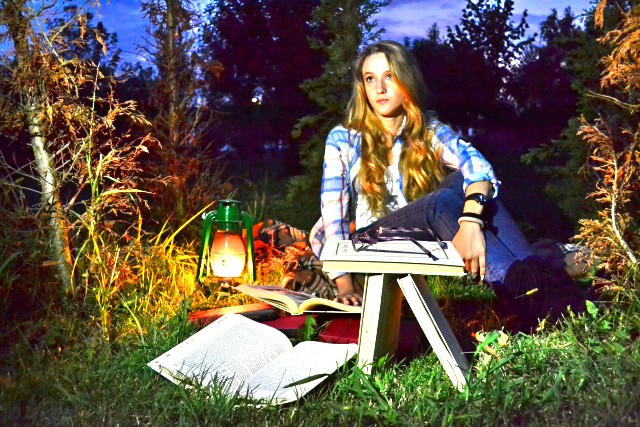} 
 \end{minipage}
 \hfill
 
 \vspace*{-\vsp}
 (d) LIME~\cite{Guo_LIME}

 \vspace*{\vsp} \vspace*{\vsp}

\hfill
 \begin{minipage}[t]{\miniwidth}
  \center
  \includegraphics[width=\imgwidth]{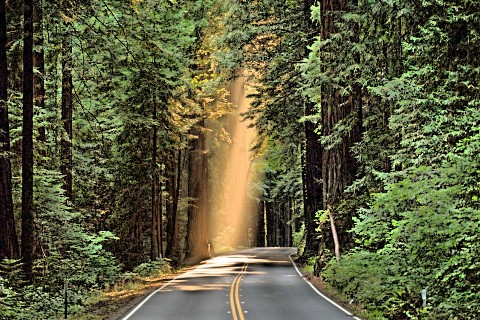}
 \end{minipage}
 \hfill
 \begin{minipage}[t]{\miniwidth}
  \center 
  \includegraphics[width=\imgwidth]{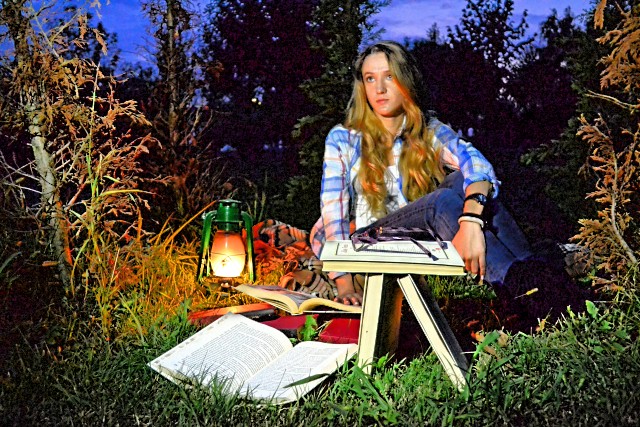} 
 \end{minipage}
 \hfill
 
 \vspace*{-\vsp}
 (e) Proposed algorithm

 \vspace*{-1mm}
 \caption{Examples of low-light image enhancement results.}
 \label{fig:results}

 \vspace*{-2.75mm}
\end{figure}

Examples of low-light image enhancement results are shown in Fig.~\ref{fig:results} \footnote{The code is available on http://www.ok.sc.e.titech.ac.jp/res/IC/LowLight/}
We picked up several low-light images from the web page~\cite{pixabay}.
Here, we compare the proposed algorithm with histogram equalization by photoshop, Matlab HDR (High-Dynamic Range) image tone mapping, and LIME~\cite{Guo_LIME}.
We simply applied the HDR tone mapping to enhance the dark region, although the input image is normal 8-bit image, or the LDR (Low-Dynamic Range) image.
For comparison algorithms, we applied default parameters.
For the proposed algorithm, the parameters $\beta$ and $\tau$ are set 15 and 50 for 8-bits image.
Those comparisons demonstrate that the proposed algorithm can effectively enhance the low-light images. 
For example, the results of LIME~\cite{Guo_LIME} in Fig.~\ref{fig:results}-(d) include some saturated regions like sunlight in left image and text of book in right image, while the results of the proposed algorithm in Fig.~\ref{fig:results}-(e) can enhance the images without such saturated region.

\section{Conclusion}
We have proposed the gradient-based low-light image enhancement algorithm.
The key idea is to enhance the gradient of the dark-region.
In addition, we have incorporated the intensity-range constraint for the image integration with the enhanced gradients. 
The experiments demonstrate our simple proposed approach can effectively enhance the low-light images.
Our future works include adaptive parameter tuning.

A part of this study was supported by ImPACT Program of Council for Science, Technology and Innovation (Cabinet Office, Government of Japan).




\begin{thebibliography}{1}

\bibitem{Russ_HistEq}
Russ, J.C., Image Processing HandBook (Third Edition), CRC Press, 1998. 

\bibitem{Yuan_Exposure}
L. Yuan and J. Sun, Automatic exposure correction of consumer photographs, European Conference on Computer Vision (ECCV), pp. 771-785, 2012.

\bibitem{Guo_LIME}
X. Guo, Y. Li, and H. Ling, LIME: Low-light image enhancement via illumination map estimation, IEEE Transactions on Image Processing, Vol. 26, No. 2, pp. 982-993, 2017.

\bibitem{Li_LowLight}
L. Li, R. Wang, W. Wang, and W. Gao, A low-light image enhancement method for both denoising and contrast enlarging, IEEE International Conference on Image Processing, pp. 3730-3734, 2015.

\bibitem{Agrawal_GradientCourse}
A. Agrawal and R. Rasker, Gradient domain manipulation techniques in vision and graphics, ICCV course, 2017.

\bibitem{Bhat_GradientShop}
P. Bhat, C. L. Zitnick, M. Cohen, and B. Curless, Gradientshop: A gradient-domain optimization framework for image and video filtering. ACM Transactions on Graphics (TOG), Vol. 29, No. 2, 2010.

\bibitem{Shibata_IntensityRange}
T. Shibata, M. Tanaka, and M. Okutomi, Gradient-domain image reconstruction framework with intensity-range and base-structure constraints, IEEE Conference on Computer Vision and Pattern Recognition (CVPR), pp. 2745-2753, 2016.

\bibitem{Tanaka_SiggraphPoster}
M. Tanaka, R. Kamio and M. Okutomi, Seamless Image Cloning by a Closed Form Solution of a Modified Poisson Problem, ACM SIGGRAPH Conference and Exhibition on Computer Graphics and Interactive Techniques in Asia (SIGGRAPH Asia Poster), p.c08-0164-1-1, 2012

\bibitem{Shibata_EI}
T. Shibata, M. Tanaka and M. Okutomi, Multi-spectrum to RGB with Direct Structure-tensor Reconstruction, IS\&T/SPIE Electronic Imaging (EI2016), 2016.

\bibitem{pixabay}
http://www.pixabay.com/

\end{thebibliography}
%

\end{document}